%% file: paper.tex
\documentclass[]{article}
\usepackage{psfrag,graphicx}
\begin{document}

\title{Safe Cooperative Robot Dynamics on Graphs\thanks{Supported 
	in part by National Science Foundation Grant IRI-9510673 [DK]
	and by National Science Foundation Grant DMS-9971629 [RG].
	A sketch of these ideas appeared as a conference proceedings
	\cite{GK:japan}.}}

\author{Robert W. Ghrist\thanks{
 School of Mathematics,
 Georgia Institute of Technology,
 Atlanta, GA 30017, USA. {\tt ghrist@math.gatech.edu}}
\and
 Daniel E. Koditschek\thanks{
 Department of Electrical Engineering and Computer Science,
 The University of Michigan,
 Ann Arbor, MI 48109-2110, USA. {\tt kod@eecs.umich.edu}}
}

\maketitle

\input setups.tex

%%%%%%%%%%%%%%%%%%%%%%%%%%%%%%%%%%%%%%%%%%%%%%%%%%%%%%%%%%%%%%%%%%%

\begin{abstract}

This paper initiates the use of vector fields to design, optimize, 
and implement reactive schedules for safe cooperative robot 
patterns on planar graphs.  
We consider Automated Guided Vehicles (AGV's) operating upon 
a predefined network of pathways. In contrast to the case of
locally Euclidean configuration spaces, regularization of 
collisions is no longer a local procedure, and issues concerning
the global topology of configuration spaces must be addressed.
The focus of the present inquiry is the achievement of safe, efficient, 
cooperative patterns in the simplest nontrivial example
(a pair of robots on a Y-network) by means of a state-event 
heirarchical controller.

\end{abstract}

%\begin{keywords} 
%configuration spaces, AGV, graph network
%\end{keywords}

%\begin{AMS}
%primary: 93C85, 68T40; secondary: 93C25, 37N35
%\end{AMS}

\pagestyle{myheadings}
\thispagestyle{plain}
\markboth{ROBERT W. GHRIST AND DANIEL E. KODITSCHEK}{SAFE ROBOT DYNAMICS}

%%%%%%%%%%%%%%%%%%%%%%%%%%%%%%%%%%%%%%%%%%%%%%%%%%%%%%

%\input{sec_1.tex}
\section{Introduction}
\label{sec_1}

Recent literature suggests the growing awareness of a need for
``reactive'' scheduling wherein one desires not merely a single
deployment of resources but a plan for successive re-deployments
against a changing environment \cite{smith.reactive}.  But
scheduling problems have been traditionally solved by appeal to a
discrete representation of the domain at hand.  Thus the need for
``tracking'' changing goals introduces a conceptual dilemma: there is
no obvious topology by which proximity to the target of a given
deployment can be measured.  In contrast to problems entailing the
management of information alone, problems in many robotics and
automation settings involve the management of { \em work } --- the
exchange of energy in the presence of geometric constraints.  In these
settings, it may be desirable to postpone the imposition of a
discrete representation long enough to gain the benefit of the natural
topology that accompanies the original domain.

This paper explores the use of vector fields for 
reactive scheduling of safe cooperative robot patterns
on graphs.  The word ``safe'' means that obstacles --- designated
illegal portions of the configuration space --- are avoided.  The word
``cooperative'' connotes situations wherein physically distributed
agents are collectively responsible for executing the schedule.  The
word ``pattern'' refers to tasks that cannot be encoded simply in terms
of a point goal in the configuration space.  The word ``reactive''
will be interpreted as requiring that the desired pattern
reject perturbations:  conditions close but slightly removed from
those desired remain close and, indeed, converge toward the exactly
desired pattern.

\subsection{Setting: AGV's on a Guidepath Network of Wires}

An automated guided vehicle (AGV) is an unmanned powered cart
``capable of following an external guidance signal to deliver a unit
load from destination to destination'' where, in most common
applications, the guidepath signal is buried in the floor
\cite{agvhandbook}.  Thus, the AGV's workspace is a network of wires
--- a graph.
The motivation to choose AGV based materials handling systems over
more conventional fixed conveyors rests not simply in their ease of
reconfigurability but in the potential they offer for graceful
response to perturbations in normal plant operation.  In real
production facilities, the flow of work in process fluctuates
constantly in the face of unanticipated workstation downtime,
variations in process rate, and, indeed, variations in materials
transport and delivery rates \cite{gershwin}.
Of course, realizing their potential robustness against these
fluctuations in work flow remains an only
partially fulfilled goal of contemporary AGV systems.

Choreographing the interacting routes of multiple AGVs in a
non-conflicting manner presents a novel, complicated, and necessarily
on-line planning problem.  Nominal routes might be designed offline
but they can never truly be traversed with the nominal timing, for all
the reasons described above.  Even under normal operating conditions,
no single nominal schedule can suffice to coordinate the workflow as
the production volume or product mix changes over time: new vehicles
need to be added or deleted and the routing scheme adapted.
In any case, abnormal conditions --- unscheduled process down times; blocked
work stations; failed vehicles --- continually arise, demanding  
altered routes.

The traffic control schemes deployed in
contemporary AGV systems are designed  to  simplify the real-time
route planning and  adaptation process by  ``blocking zone control''
strategies.
The workspace is partitioned into a small number of cells and,
regardless of the details of their source and destination tasks, no two
AGVs are
ever allowed into the same cell at the same time \cite{agvhandbook}.
Clearly, this simplification results in significant loss of a
network's traffic capacity.

In this paper, we will consider a centralized approach that employs
dynamical systems theory to focus on real-time responsiveness and
efficiency as opposed to computational complexity or average
throughput.  No doubt, beyond a certain maximum number of vehicles,
 the necessity to compute in the high dimensional configuration
space will limit the applicability of any algorithms that arise.
However, this point of view seems not to have been carefully explored
in the literature.  Indeed, we will sketch some ideas about how an
approach that starts from the coupled version of the problem may lend
 sufficient insight to move back and forth between the individuals'
and the group's configuration spaces even in real time.  For the sake
of concreteness we will work in the so-called ``pickup and delivery''
(as opposed to the ``stop and go'' \cite{bozer.iie91}) paradigm of
assembly or fabrication, and we will not be concerned with
warehousing style  AGV applications.

\subsection{Contributions of the Paper}

The paper is organized as follows.  In \S\ref{sec_2}, we review
fundamental facts about the topology of graphs, with which we
define the class of {\em edge point
fields} ---  locally defined dynamics that realize 
single letter patterns. These  act collectively as a toolbox 
from which to build a hybrid controller for achieving arbitrary 
patterns with a single AGV. This represents a slight generalization of
the scheme the second author and colleagues have proposed in
\cite{kod&burridge.ijrr}. 

The problem of dynamics and control on non-trivial graphs is then
considered in \S\ref{sec_3}, beginning with a detailed discussion of
a natural intrinsic coordinate system in which to frame the
configuration space. We present a topological analysis of
the configuration space for a pair of AGV's on a Y-shaped graph ---
the simplest nontrivial situation. Here, a clarification of the 
configuration space presentation leads easily to a vector field 
construction that brings all initial conditions of two robots
on the graph to any desired pair of goal points while guaranteeing
safety (i.e., no collisions along the way). The desire
for a more decoupled controller --- the hope of an ``interleaving'' of
otherwise independent individual  patterns --- impels  a revised approach to
safe navigation leading to the construction of a vector field that enables
the AGV's to ``dance'' about one other at a vertex.

The dynamical features of this {\em circulating field} are suggestive
of a hybrid construction that would allow multiple independent
patterns to be safely interleaved. We proceed in \S\ref{sec_4} to 
construct a 12-symbol grammar of so-called ``monotone'' cycles: those
patterns which exclude multiple robots on a single edge of the graph. 
The goal of this excursion is to tune limit cycles visiting 
various docking stations in such a way as to be optimal with 
respect to certain notions of distance or performance.

We complete our treatment of this fundamental example by 
synthesizing the grammar results into a state-event controlled hybrid
system for achieving cooperative patterns.
Appendix A is included to place on a rigorous foundation 
the use of vector fields on graphs and configuration spaces
thereof.

\section{Notation and Background}
\label{sec_2}

\subsection{Graph Topology}

A {\em graph}, $\Graph$, consists of a finite collection of 
0-dimensional vertices $\Vs \ldf \{ v_{i}\}_1^N$, 
and 1-dimensional edges $\Es \ldf \{ e_{j} \}_1^M$ assembled
as follows. Each edge is homeomorphic to the closed interval
$[0,1]$ attached to $\Vs$ along its boundary points $\{0\}$ 
and $\{1\}$.\footnote{
We will assume away in the sequel the possibility of ``homoclinic''
edges whose  boundary points are attached to the 
same vertex.}
We place upon $\Graph$ the quotient topology
given by the endpoint identifications:
Neighborhoods of a point in the interior of $e_j$ are homeomorphic
images of interval neighborhoods of the corresponding point in $[0,1]$, 
and neighborhoods of a vertex $v_i$ consist of the union 
of homeomorphic images of half-open neighborhoods of the 
endpoints for all incident edges.

The configuration spaces we consider in \S\ref{sec_3} and following 
are self-products of graphs. The topology of $\Graph\times\Graph$ 
is easily understood in terms of the topology of $\Graph$ as 
follows \cite{munkres}. Let $(x,y)\in\Graph\times\Graph$ denote
an ordered pair in the product. Then any small neighborhood of
$(x,y)$ within $\Graph\times\Graph$ is the union of neighborhoods 
of the form $\Ns(u)\times \Ns(v)$, where $\Ns(\cdot)$
denotes neighborhood within $\Graph$. In other words, the 
products of neighborhoods form a {\em basis} of neighborhoods
in the product space. 

Given a graph, $\Graph$, outfitted with a finite number $N$ of 
non-colliding AGV's constrained to move on $\Graph$, the
(labeled) configuration space of safe motions is defined as
\be
	\Cs\ldf \left(\Graph\times\ldots\times\Graph\right) - 
	\Ns(\Diag),
\ee
where $\Diag\ldf\{(x_i)\in\Graph\times\ldots\times\Graph :
x_j = x_k$ for some $j\neq k\}$ denotes the pairwise
diagonal and $\Ns(\cdot)$ denotes (small) neighborhood. 

For general graphs, the topological features of $\Cs$ can be extremely
complicated.
We do not treat the general aspects of this problem comprehensively in
this paper; rather, we restrict attention to the simplest nontrivial 
example which illustrates nicely the relevant features present in the 
more general situation. The topological characteristics 
of general configuration spaces
on graphs is treated in \cite{Ghr:birman,Ghr:homotopytype}. 
Mathematically, it is usually most interesting to pass to 
the quotient of $\Cs$ by the action of the permutation group
on $N$ elements, thus forgetting the identities of the AGV 
elements; however, as such spaces are almost completely divorced
from any applications involving coordinated transport, we work on the 
``full'' configuration space $\Cs$.

In order to proceed, it is necessary to clarify what we mean by a
vector field on a simplicial complex that fails to be a manifold. This
is a nontrivial issue: for example, in the case of a graph, the
tangent space to a vertex with incidence number greater than two is
not well-defined.
We defer a more detailed discussion of these statements to 
Appendix A. The essential difference is that we construct
{\em semiflows}: flows which possess unique forward orbits.

% -----------------------------------------------------------
 \subsection{Edge Point Fields for Single AGV Control}
% -----------------------------------------------------------

In the context of describing and executing {\em patterns} or periodic
motions on a graph, one desires a set of building blocks for moving
from one goal to the next. We introduce the terminology and 
philosophy for constructing patterns by way of the simplest 
possible examples: a single AGV on a graph. This avoids the 
additional topological complications present in the context of 
cooperative motion. 

We thus introduce the class of {\em edge 
point fields} as a dynamical toolbox for a hybrid controller. 
Given a specified goal point $\goal\in e_{j}$ within an edge of
$\Graph$, an {\em edge point field} is a locally defined vector field 
$\EPF_\goal$ on $\Graph$ with the following properties:
\begin{description}

        \item[Locally Defined:] $\EPF_\goal$ is defined on a
	neighborhood $\Ns(e_j)$ of the goal-edge $e_j$ within the graph 
	topology. Furthermore, forward orbits under $\EPF_\goal$ 
	are uniquely defined.

        \item[Point Attractor:]  every forward orbit of $\EPF_\goal$ 
	asymptotically approaches the unique fixed point 
	$\goal\in e_j$.\footnote{
	When it is not clear from the context, we shall denote the
	goal point achieved by an edge point flow as
        $\Goal(\EPF_\goal) = \{ \goal \}.$}

        \item[Navigation-Like:] $X_\goal$ admits a $C^0$ Lyapunov 
	function, $\obj_\goal:\Graph\rightarrow\real$.

\end{description}

The following existence lemma (whose trivial proof we omit) holds.
\llem{EPF}
{
Given any edge $e_j\subset\Graph$ which is contractible within
$\Graph$, there exists an edge point field $\EPF_\goal$ for 
any desired goal $\goal\in e_j$.
}\label{lem_EdgePoint}

The only occasion for which an edge $e_j$ is not contractible in
$\Graph$ is in the ``homoclinic case'' when both endpoints of $e_j$
are attached to the same vertex, forming a loop. In such instances,
one may avoid the problem by subdividing the edge to include more
vertices, which is very natural in the setting of this paper, since
vertices correspond to workstations along a path.

% --------------------------------------------------
\subsection{Discrete Regulation of Patterns}
% --------------------------------------------------

We adopt the standard framework of symbolic dynamics
\cite{lind&marcus}.  By an excursion on a graph is meant a (possibly
infinite) sequence of edges from the graph, 
$E=e_{i_{1}}\ldots e_{i_{N}}\ldots \in \Es^{Z}$,
having the property that each pair of contiguous edges, $e_{i_{j}}$ 
and $e_{i_{j+1}}$ share a vertex in common.
The set of excursions forms a language, $\Ls$: the
so-called {\em subshift} on the alphabet defined by the named edges
(we assume each name is unique) \cite{lind&marcus}.  The {\em shift
operator}, $\sigma$, defines a discrete dynamical system on the set of
excursions, mapping the set of infinite sequences into itself by
decrementing the time index.  An {\em $M$-block extension} of the 
original language arises in the obvious way from grouping together each
successive block of $M$ contiguous letters from an original sequence,
and it is clear how $\sigma$ induces a shift operator, $\sigma^{M}$
on this derived set of sequences.

Given a legal block, $B=e_{i_{1}}\ldots e_{i_{M}}\in \Ls$, 
we will say that an excursion realizes that pattern if its $M$-block
extension eventually reaches the ``goal'' $BBBBB \ldots $ under the
iterates of $\sigma^{M}$. In other words, after some finite number of
applications of $\sigma$,  the excursion consists of repetitions
of the block $B$ (terminating possibly with the empty edge).

In a previous paper \cite{kod&burridge.ijrr}, the second author and
colleagues introduced a very simple but effective discrete event
controller for regulating patterns on graphs from all reachable
initial edges by pruning the graph back to a tree (imposing an
ordering).  Of course, this simple idea has a much longer history. In
robotics it was introduced in \cite{lmt} as ``pre-image
backchaining;'' pursued in \cite{mason.funnel} as a method for
building verifiable hardened automation via the metaphor of a
funneling; and in \cite{erdmann.sensors} as a means of prescribing
sensor specifications from goals and action sets.  In the discrete
event systems literature an optimal version of this procedure has been
introduced in \cite{heyman.is93} and a generalization recently has
been proposed in
\cite{stephane.opt}.

 Let $\Es^{0} 
\ldf B \subset \Es $ denote the edges of $\Graph$ that appear in the
block of letters specifying the desired  pattern. Denote by
\[
\Es^{n+1} \subset \Es -  \bigcup_{k\leq n}\Es^{k}
\]
those edges that share a vertex with an edge in $\Es^{n}$ but are not
in any of the previously defined subsets.  This yields a finite
partition of $\Es$ into ``levels,'' $\{ \Es^{p} \}_{p=0}^{P}$,
such that for each edge, $e^p_i \in \Es^{p}$, there can be found a
legal successor edge, $e^{p-1}_j \in \Es^{p-1}$, such that
$e^p_ie^{p-1}_j \in \Ls$ is a legal block in the language.  Note
that we have implicitly assumed $\Es^{0}$ is reachable from the
entire graph --- otherwise, there will be some ``leftover'' component
of $\Es$ forming the last cell in the partition starting within which
it is not possible to achieve the pattern. Note as well that we impose
some ordering of each cell $\Es^{p} = \{ e^p_i \}_{i=1}^{M_{p}}$: 
the edges of $\Es^{0}=B$ are ordered by their appearance in the block; 
the ordering of edges in higher level cells is arbitrary.

We may now define a ``graph controller'' law, $G \st \Es
\rightarrow \Es$ as follows. From the nature of the partition 
$\{\Es^p\}$ above, it is clear that the least legal successor function,
\beq{successor}
{
\choicetwo{L(e_i^p)}
{i+1 \: {\mbox{mod}} \: M}{\, p=0}
{\min \{ j \leq M_p \,\st\, e^p_ie^{p-1}_j \in \Ls \}}{\, p>0} ,
}
is well-defined. From this, we construct the graph controller:
\beq{graph_control}
{
G(e^p_i) \ldf e^{p-1}_{L(p,i)} .
}
It follows almost directly from the definition of this function that
its successive application to any edge leads eventually to a
repetition of the desired pattern:
\lprop{successor}
{
The iterates of $G$ on $\Es$ achieve the pattern $B$.
}

\subsection{Hybrid Edge Point Fields}

A semiflow, $(X)^t$, on the graph induces excursions in $\Ls$  
parametrized by an initial condition as follows.  The
first letter corresponds to the edge in which the initial condition is
located (initial conditions at vertices are assigned to the
incident edge along which the semiflow points).  
The next letter is added to the sequence by motion
through a vertex from one edge to the next.

We will say of two edge point fields, $\EPF_{1},\EPF_{2}$ on a graph,
$\Graph$, that $\EPF_{1}$ { \em prepares } $\EPF_{2}$, denoted 
$\EPF_{1} \succ \EPF_{2},$ if the goal of the first is in the 
domain of attraction of the second,
$ \Goal(\EPF_{1}) \subset \Ns(\EPF_{2}).  $
Given any finite collection of edge point fields on
$\Graph$, we will choose some $0 < \alpha < 1$ and assume that their 
associated Lyapunov functions have been scaled in such a
fashion that $\EPF_{1} \succ  \EPF_{2},$ implies
$
\inv{\prl{\obj_{1}}}[0,\alpha] \subset \Ns(\EPF_{2}).
$
In other words, an $\alpha$ crossing of the trajectory
$\obj_{1} \circ \prl{\EPF_{1}}^{t}$
signals arrival in  $\Ns(\EPF_{2})$.

Suppose now that for every edge in some pattern block, $e^0_i \in
\Es^{0}$, there has been designated a goal point, $\goal^0_i$, 
along with an edge point field $\EPF^0_i$ taking that goal, 
$\Goal(\EPF^0_i)=\goal^0_i$.
Assume as well that the edge point field  associated with  each previous
edge in the pattern prepares the flow associated with the next edge,
in other words, using the successor function \req{successor} we have,
\[
\Goal \prl{ \EPF^0_{j} } \subset \Ns\prl{\EPF^0_{L(j)}}.
\]
Now construct edge point fields on all the edges of $\Graph$ such that
the tree representation of their $\succ$ relations is exactly the tree
pruned from the original graph above ---  namely we have
\[
\Goal \prl{ \EPF^p_{j} } \subset \Ns \prl{\EPF^{p-1}_{L(j)}}.
\]
% ****************************************************

We are finally in a position to construct a hybrid semi-flow on
$\Graph$. This feedback controller will run the piece-wise
smooth vector field, $\dot x = \EPF$, as follows
\beq{hybrid}
{
\choicetwo{\EPF}
{\EPF^p_j}{ x \in e^p_j {\mbox{ and }} \obj^p_j > \alpha }
{\EPF^{p-1}_{L(j)} }{x \in e^{p-1}_{L(j)} {\mbox{ or }} \obj^p_j \leq 
	\alpha } .
}
It is clear from the construction that progress from edge to edge
of the state of this flow echoes the graph transition rule $G$,
constructed above.

\lprop{hybrid}
{The edge transitions induced by the hybrid controller \req{hybrid}
are precisely the iterates of the graph map, $G$,
\req{graph_control} in the language, $\Ls$.
}

\section{The Y-Graph}
\label{sec_3}
% *********************************************************

We now turn our attention to the safe control of multiple AGV's 
on a graph work-space via vector fields. 
For the remainder of this work, we consider the simplest example
of a non-trivial configuration space: that associated to 
the {\em Y-graph}, $\Ygraph$, having four vertices $\{v_i\}_0^3$
and three edges $\{e_i\}_1^3$, as illustrated in Figure~\ref{fig_Y}. 
Although this is a simple scenario compared to what one 
finds in a typical setting, there are several 
reasons why this example is in many respects canonical. 
\begin{enumerate}
\item	{\bf Simplicity:} 
	Any graph may be constructed by gluing 
	$K$-prong graphs together for various $K$. The 
	$K=3$ model we consider is the simplest nontrivial
	case and is instructive for understanding the richness
	and challenges of local cooperative dynamics on graphs.
\item 	{\bf Genericity with respect to graphs:}
	Graphs which consist of copies of $\Ygraph$ glued
	together, the {\em trivalent graphs}, are generic in the
	sense that any nontrivial graph may be perturbed 
	in a neighborhood of the vertex set so as to be trivalent.
	For example, the 4-valent graph resembling the letter
	`X' may be perturbed slightly to resemble the letter
	`H' --- a trivalent graph. An induction argument shows
	that this is true for all graphs. Hence, the dynamics 
	on an arbitrary graph are approximated by patching
	together dynamics on copies of $\Ygraph$.
\item	{\bf Genericity with respect to local dynamics:} 
	Finally, pairwise local AGV interactions on an arbitrary
	graph restrict precisely to the dynamics of two agents on
	$\Ygraph$ as follows. Given a vertex $v$ of a graph $\Gamma$,
	assume that two AGV's, $x$ and $y$, are on different edges
	$e_1$ and $e_2$ incident to $v$ and moving towards 
	$v$ with the goal of switching positions. A collision is 
	imminent unless one AGV ``moves out of the way'' onto
	some other edge $e_3$ incident to $v$. The local interactions
	thus restrict to dynamics of a pair of AGV's on the subgraph 
	defined by $\{v;e_1,e_2,e_3\}$. Hence, the case we treat in
	this paper is the generic scenario for the local resolution 
	of collision singularities in cooperative dynamics on graphs.
\end{enumerate}

% *********************************************************
\subsection{Intrinsic Coordinates}
% *********************************************************

The configuration space of $N$ points on $\Ygraph$ is a 
subset of the $N$-fold cartesian product 
$\Ygraph\times\Ygraph\times\cdots\times\Ygraph$. Since each 
graph which is physically relevant to the setting of this paper
is embedded in a factory floor or ceiling and thus planar, 
the configuration space $\Cs^N(\Ygraph)$ embeds 
naturally in $R^{2N}$. We wish to modify this embedding 
to facilitate both analysis on and visualization of the configuration
space. We will present alternate embeddings in both higher and 
lower dimensional Euclidean spaces for these purposes.

We begin with representing the configuration space within a 
higher-dimensional Euclidean space via a coordinate system 
which is {\em intrinsic}: it is independent 
of how the graph is embedded in space. We illustrate this 
coordinate system with the Y-graph $\Ygraph$, noting that 
a few simple modifications yields a coordinate scheme for 
more general graphs. 

Let $\{e_i\}_1^3$ denote the three edges in $\Ygraph$, parametrized
so that $e_i$ is identified with $[0,1]$ with each $\{0\}$ at the 
center $v_0$ of $\Ygraph$. Any point in 
$\Ygraph$ is thus given by a vector $x$ in the
$\{e_i\}$ basis whose magnitude $\Value{x}\in[0,1]$ determines the 
position of the point in the $e_i$ direction. Denote by 
$\Index{x}$ the value of $i$ so that $x=\Value{x}e_{\Index{x}}$.
This parametrization embeds $\Ygraph$
as the positive unit axis frame in $R^3$.
Likewise, a point in $\Cs$ is given as a pair of distinct vectors
$(x,y)$, i.e., as the positive unit axis frame in $R^3$ cross itself
sitting inside of $R^3\times R^3\cong R^6$. We have thus embedded
the configuration space of two distinct points on $\Ygraph$ 
in the positive orthant of $R^6$. It is clear that one can embed the
more general configuration space of $N$ points on $\Ygraph$ 
in $R^{3N}$ in this manner.

This coordinate system is particularly well-suited to describing 
vector fields on $\Cs$ and in implementing numerical simulations
of dynamics, as the coordinates explicitly keep track of the 
physical position of each point on the graph. 

% *********************************************************
\subsection{A Topological Analysis}
% *********************************************************

More useful for visualization purposes, however, is the following 
construction which embeds $\Cs$ within $R^3$.

% =========================================
\begin{figure}[htb]
\begin{center}
\psfragscanon
\psfrag{0}[][]{$v_0$}
\psfrag{1}[][]{$v_1$}
\psfrag{2}[][]{$v_2$}
\psfrag{3}[][]{$v_3$}
\psfrag{a}[][]{$e_1$}
\psfrag{b}[][]{$e_2$}
\psfrag{c}[][]{$e_3$}
	{%\centerline{
	\includegraphics[width=1.6in]{{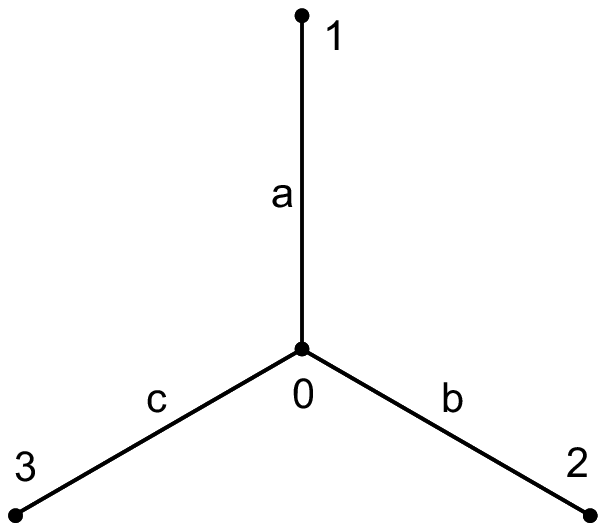}}
\hspace{0.2in}
	\includegraphics[width=3.0in]{{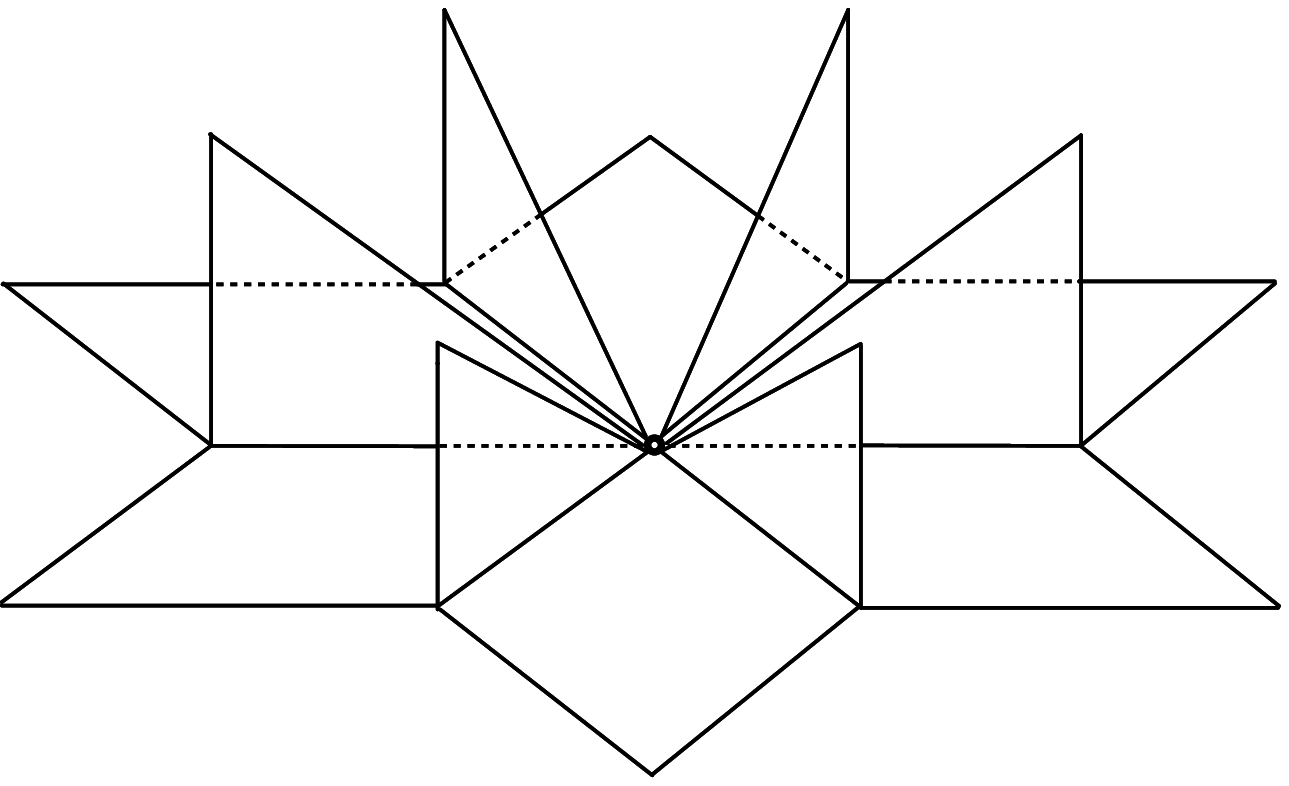}}}
\end{center}
\vspace{-0.1in}

\caption{[left] The $Y$-graph $\Upsilon$; [right] the 
configuration space $\Cs$ embedded in $R^3$.} 
\label{fig_pin5}\label{fig_Y}
\end{figure}
% =========================================

\begin{thm}\label{thm_Space}
The configuration space $\Cs$ associated to a pair of AGV's 
restricted to the $Y$-graph $\Ygraph$ is homeomorphic to 
a punctured disc with six 2-simplices attached as in 
Figure~\ref{fig_pin5}.
\end{thm}

\begpf
Recall that $\Cs$ consists of pairs of distinct vectors $(x,y)$ in intrinsic 
coordinates. Restrict attention to the subspace $\Finless\subset\Cs$ 
defined by 
\be
	\Finless := \{(x,y)\in\Cs : \Index{x}\neq\Index{y}\} ,
\ee 
where an undefined index is considered to be different than
one which is defined. Thus, $\Finless$ consists of those configurations 
for which both AGV's do not occupy the same edge interior.

The set $\Finless$ has a natural cellular decomposition 
as follows. There are $2$ AGV's and $3$ edges in $\Ygraph$; hence,
there are $3\cdot 2 = 6$ cells $\Finless_{i,j}\subset\Finless$ where
$i:=\Index{x}\neq\Index{y}=:j$. Since (the closure of) 
each edge in $\Ygraph$ is homeomorphic to $[0,1]$ (determined by 
$\Value{\cdot}$), the cell $\Finless_{i,j}$ is homeomorphic to 
$([0,1]\times[0,1])-\{(0,0)\}$, where, of course, the origin $(0,0)$ 
is removed as it belongs to the diagonal $\Diag$. 
A path in $\Finless$ can move from cell to cell only along the subsets 
where the index of one AGV changes: e.g., $\Value{x}=0$ or 
$\Value{y}=0$. Thus, the edges $\{0\}\times(0,1]$ and 
$(0,1]\times\{0\}$ of the punctured square $\Finless_{i,j}$ 
are attached respectively to $\Finless_{k,j}$ and $\Finless_{i,k}$,
where $k$ is the unique index not equal to $i$ or $j$.

Furthermore, each 2-cell $\Finless_{i,j}$ has a product structure
as follows: decompose $\Finless_{i,j}$ along lines of constant 
$\theta := \tan^{-1}\left(\frac{\Value{y}}{\Value{x}}\right)$.
It is clear that $\theta$ is the angle in the unit first quadrant in 
which $\Finless_{i,j}$ sits. Hence, each $\Finless_{i,j}$ is decomposed
into a product of a closed interval $S_{i,j}:=\theta\in[0,\pi/2]$ 
(an `angular' coordinate) with the 
half-open interval $(0,1]$ (a `radial' coordinate). As this product
decomposition is respected along the gluing edges, we have a decomposition
of all of $\Finless$ into the product of $(0,1]\times S$, where $S$
is a cellular complex given by gluing the six segments $S_{i,j}$ 
end-to-end cyclically along their endpoints. 
The set $S$ is a 1-manifold without boundary since each $S_{i,j}$ 
is a closed interval, each of whose endpoints is glued to precisely one 
other $S_{i,j}$. Hence, by the classification of 1-manifolds, 
$S$ is homeomorphic to a circle. We have thus decomposed 
$\Finless$ as the cross product of a circle with $(0,1]$ --- 
a punctured unit disc. 

The complement of $\Finless$ in $\Cs$ consists of those regions 
where $\Index{x}=\Index{y}$. For each $i=1..3$, the subset of 
$\Cs$ where $\Index{x}=\Index{y}=i$ is homeomorphic to 
$((0,1]\times(0,1])-\{\Value{x}=\Value{y}\}$: this consists of 
two disjoint triangular ``fins.'' A total of six such fins are thus 
attached to $\Finless$ along the six edges where $\Value{x}$ or
$\Value{y}=0$. In the coordinates of the product decomposition 
for $\Finless$, these fins emanate along the radial lines where 
$\theta$ equals zero or $\pi/2$, yielding the topological space 
illustrated in Figure~\ref{fig_pin5}.
\endpf

The precise drawing of Figure~\ref{fig_pin5} represents this 
punctured disc $\Finless$ as a hexagon-shaped complex with a
punctured center: this follows from the cellular structure of 
$\Finless$ as being built from six squares sewn together. 

\lcor{sphereworld}
{
Given any pair of goals $\goal\ldf(\goal_1,\goal_2)$ where 
$\goal_1$ and $\goal_2$ live on different branches of $\Ygraph$, 
there exists a navigation function (of class piecewise real-analytic) 
generating a semiflow which sends all but a measure-zero set of 
initial conditions to $\goal$ under the gradient semiflow. 
}

\begpf
The subset $\Finless\subset\Cs$ is homeomorphic to a punctured disc
$S\times(0,1]$, and may be easily compactified to an annulus
$S\times[0,1]$ by removing an open neighborhood of the diagonal. Then, 
the conditions for the theorems of Koditschek and Rimon
\cite{kod&rimon.aam} are met, since an annulus is a {\em sphereworld}.
Hence, not only does a navigation function $\Phi$ on this subspace exist,
but an explicit procedure for determining $\Phi$ is given 
\cite{kod&rimon.aam}. One may then 
extend $\Phi$ to the remainder of $\Cs$ as follows: choose a point 
$(x,y)$ on the fin and define 
\be
\Phi(x,y) := \left\{
\begin{array} {ccc}
\frac{\ds 1}{\ds 1-\Value{x}}\Phi(0,y) & ; & \Value{x}<\Value{y} \\
\frac{\ds 1}{\ds 1-\Value{y}}\Phi(x,0) & ; & \Value{y}<\Value{x}
\end{array} \right.,
\ee
so that $\Phi$ increases sharply along the fins. This directs the 
gradient flow to monotonically ``descend'' away from the diagonal 
and onto $\Finless$.
Note that $\Finless$ is forward-invariant under the dynamics, and, 
that upon prescribing the vector field on the fins to point 
transversally into $\Finless$, we have 
defined a semiflow, and hence a well-defined navigational procedure.
\endpf

This result is very satisfying in the sense that it guarantees a
navigation function by applying existing theory to a situation 
which, from the definition alone, would not appear to 
be remotely related to a sphereworld.  However, a deeper 
analysis of configuration spaces of graphs \cite{Ghr:homotopytype}
reveals that for more than two AGV's, the configuration 
space is never a sphereworld.\footnote{The configuration space
of a graph turns out to be {\em aspherical}: there are no
essential closed spheres of dimension larger than one.}
Hence, we consider an alternate 
solution to the problem of realizing compatible goals by means 
of a vector field on the configuration space. This 
method is adaptable to more complicated settings.

% *********************************************************
\subsection{Example: a Circulating Flow}
% *********************************************************

Before proceeding with a general scheme for controlling two agents
on the Y-graph, we present a simple example of a vector field
on the configuration space which can be used to regularize 
collisions about a generic trivalent vertex. 
Theorem~\ref{thm_Space} and Figure~\ref{fig_pin5} suggest a 
natural {\em circulating flow} on the configuration 
space $\Cs$ which has the effect of inducing a ``dance'' 
between the pair of AGV's which cycles through all combinations 
of distinct point goals.

\begin{figure}[htb]
\begin{center}
\psfragscanon
\psfrag{1}[bl][bl]{$(e_1,0)$}
\psfrag{2}[b][b]{$(0,e_3)$}
\psfrag{3}[b][b]{$(e_2,0)$}
\psfrag{4}[br][br]{$(0,e_1)$}
\psfrag{5}[b][]{$(e_3,0)$}
\psfrag{6}[b][]{$(0,e_2)$}
	{%\centerline{
	\includegraphics[width=2.1in]{{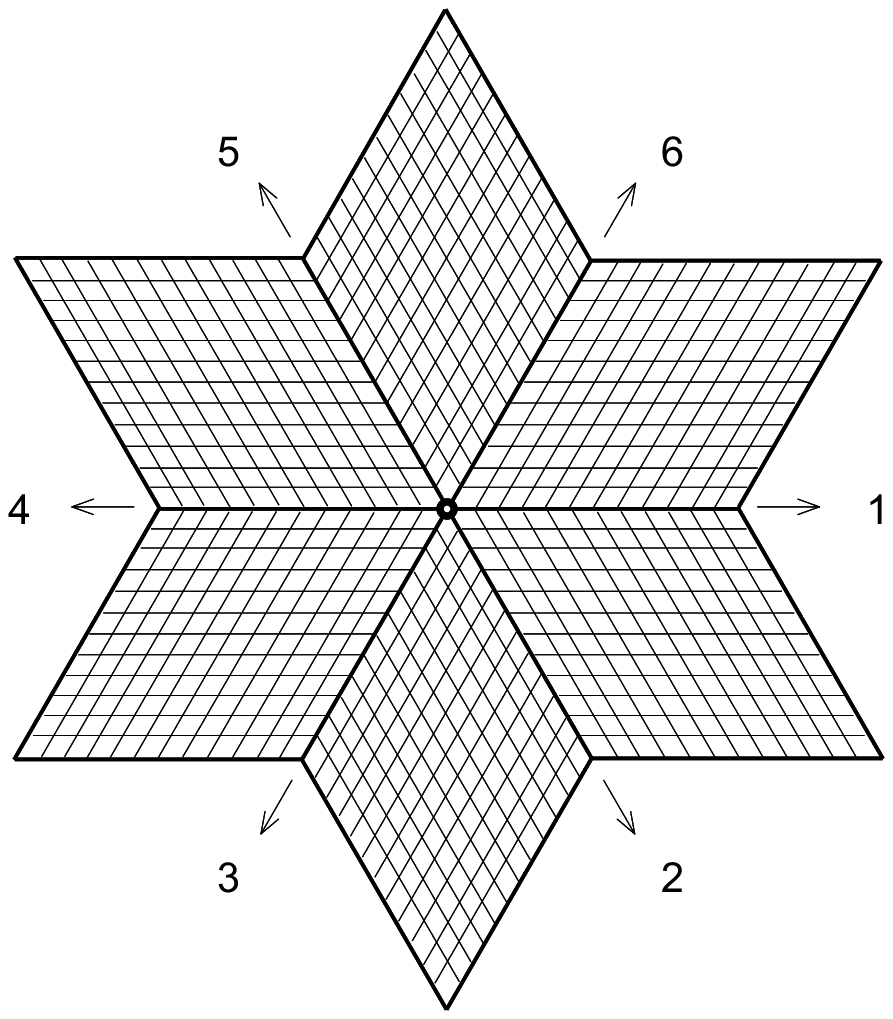}}
\hspace{0.3in}
	\includegraphics[width=2.1in]{{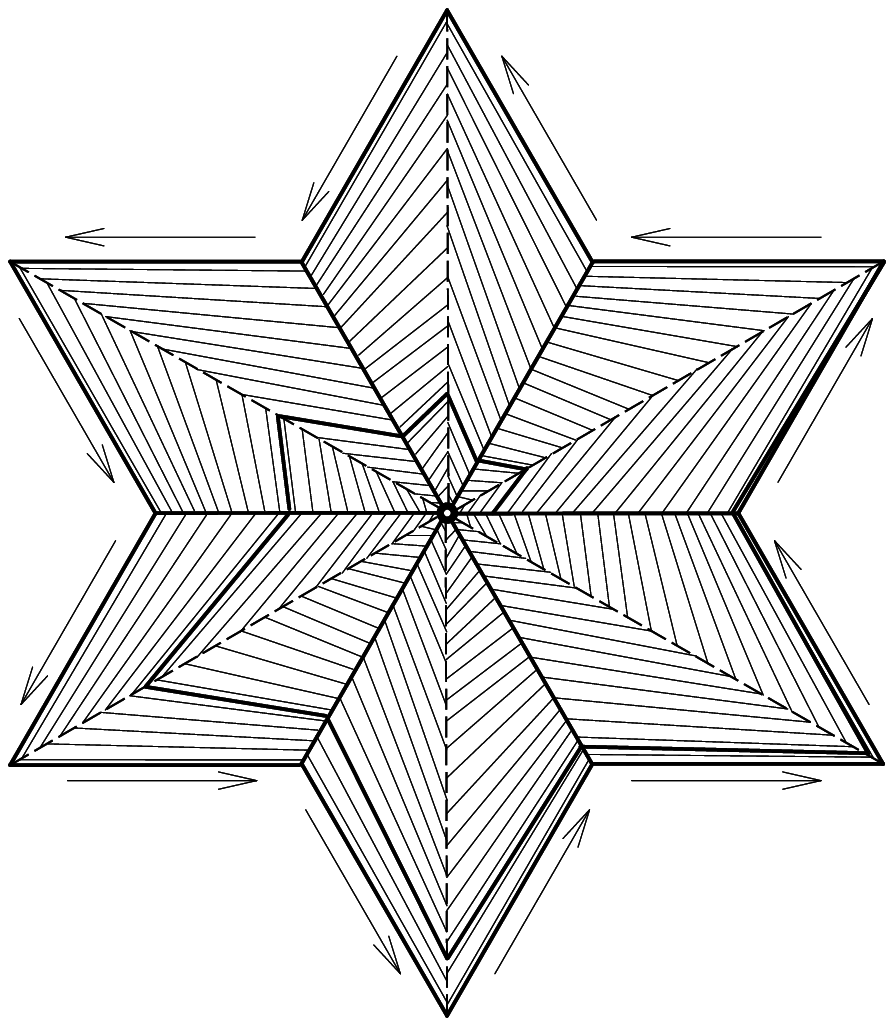}}}
\end{center}
\caption{[left] The coordinate system on the unfinned region $\Finless$ of $\Cs$; [right] The circulating flow with a typical orbit.} 
\label{fig_Coords}\label{fig_field}
\end{figure}

\begin{thm}
There exists a piecewise-smooth vector field $X$ on $\Cs$ which 
has the following properties:
\begin{enumerate}
\item $X$ defines a nonsingular semiflow on $\Cs$;
\item The diagonal $\Diag$ is repelling with respect to $X$; and
\item Every orbit of $X$ approaches a unique attracting limit cycle
	on $\Cs$ which cycles through all possible ordered pairs of 
	distinct edge-states.
\end{enumerate}
\end{thm}
\begpf
Denote by $\Finless$ that portion of the configuration space 
which corresponds to the AGV's being on distinct edges of the 
graph: as proven earlier, $\Finless$ is homeomorphic to a punctured disc. 
The intrinsic coordinates on the configuration space $\Cs$ is 
illustrated in Figure~\ref{fig_Coords}, where only $\Finless$ is shown
for illustration purposes. The reader should think of this as
a collection of six square coordinate planes, attached together
pairwise along axes with the origin removed
(this is actually an isometry for the natural product metric). The six 
triangular fins are then attached as per Figure~\ref{fig_pin5}.

% =========================================
% =========================================

Recal that any point in the graph is represented as a vector 
$x=\Value{x}e_i$ for some $i$. Denote by $\hat{e}_i$ the unit tangent
vector in each tangent space $T_xe_i$ pointing in the positive
(outward) direction towards the endpoint $v_i$. 
The vector field we propose is the following: given $(x,y)\in\Cs$, 
\begin{enumerate}
\item
	If $\Index{x}=\Index{y}$ then
\bea\label{eq_Dance1}
	\left\{\begin{array}{l}
	\dot{x} = -\Value{y}\hat{e}_{\Index{x}} \\
	\dot{y} = \Value{y}(1-\Value{y})\hat{e}_{\Index{y}}
	\end{array}\right\}
	& 0<\Value{x}<\Value{y} 
\\
	\left\{\begin{array}{l}
	\dot{x} = \Value{x}(1-\Value{x})\hat{e}_{\Index{x}} \\
	\dot{y} = -\Value{x}\hat{e}_{\Index{y}} 	
	\end{array}\right\} 
	& 0<\Value{y}<\Value{x} 
\eea
\item
	If $\Index{x} = \Index{y}+1$ or $\Value{x}=0$ then
\bea\label{eq_Dance2}
	\left\{\begin{array}{l}
	\dot{x} = \Value{y}\hat{e}_{(\Index{y}+1)} \\
	\dot{y} = \Value{y}(1-\Value{y})\hat{e}_{\Index{y}}
	\end{array}\right\}
	& 0\leq\Value{x}<\Value{y} 
\\
	\left\{\begin{array}{l}
	\dot{x} = \Value{x}(1-\Value{x})\hat{e}_{\Index{x}} \\
	\dot{y} = -\Value{x}\hat{e}_{\Index{y}} 	
	\end{array}\right\} 
	& 0<\Value{y}\leq\Value{x} 
\eea
\item
	If $\Index{y} = \Index{x}+1$ or $\Value{y}=0$ then
\bea\label{eq_Dance3}
	\left\{\begin{array}{l}
	\dot{x} = -\Value{y}\hat{e}_{\Index{x}} \\
	\dot{y} = \Value{y}(1-\Value{y})\hat{e}_{\Index{y}}
	\end{array}\right\} 
	& 0<\Value{x}\leq\Value{y} 
\\
	\left\{\begin{array}{l}
	\dot{x} = \Value{x}(1-\Value{x})\hat{e}_{\Index{x}} \\
	\dot{y} = \Value{x}\hat{e}_{(\Index{x}+1)} 	
	\end{array}\right\} 
	& 0\leq\Value{y}<\Value{x} 
\eea
\end{enumerate}
Note that all addition operations on $\Index{x}$ and $\Index{y}$
are performed mod three.

The vector field is nonsingular as follows: if $\Value{x}\Value{y}\neq 0$, 
then the vector field is by inspection nonsingular. If $\Value{x}=0$, 
then $\Value{y}>0=\Value{x}$ since the points are distinct. It then 
follows from Equation~(\ref{eq_Dance2}) that the vector field on this 
region has $d\Value{x}/dt=\Value{y}\neq 0$. A similar argument 
holds for the case where $\Value{y}=0$.

The vector field defines a semiflow as follows: on those 
regions where $0\neq\Value{x}\neq\Value{y}\neq 0$, the vector 
field is smooth and hence defines a true flow. Along the lines where
$\Value{x}=\Value{y}$, the vector field is only $C^0$, 
but nevertheless is constructed so as to define unique solution curves;
hence the region $\Finless$, where $\Index{x}\neq\Index{y}$, is invariant 
under the flow. Finally, along the branch line curves where 
$\Value{x}=0$ or $\Value{y}=0$, the vector field points into the 
the branch lines from the fins, implying that the dynamics is a semiflow
(see the remarks in Appendix A).

This vector field admits a $C^0$ Lyapunov function 
$\obj:\Cs\rightarrow[0,1)$ of the form
\be
	\obj(x,y) \ldf \left\{
	\begin{array}{rl}
	1-\left\vert (\Value{x}-\Value{y}) \right\vert & :\, 
		\Index{x}=\Index{y} \\
	1-{\mbox{max}}\left\{\Value{x},\Value{y}\right\} & :\, 
		\Index{x}\neq\Index{y}
	\end{array}\right.  .
\ee
From Equation~(\ref{eq_Dance1}), one computes that on the fins
($\Index{x}=\Index{y}$), 
\be
	\frac{d\obj}{dt} = -\abs{(\frac{d\Value{x}}{dt}-
			\frac{d\Value{y}}{dt})}<0 , 
\ee
since here $\Value{x}\neq\Value{y}$. Furthermore, on the disc $\Finless$ ($\Index{x}\neq\Index{y}$),  $\obj$ changes as 
\be
	\frac{d\obj}{dt} = \obj(\obj-1) .
\ee
Hence, $\obj$ strictly decreases off of the boundary of the disc 
\be
	\del\Finless \ldf \left\{(x,y) : \Value{x}=1 {\mbox{ or }}
		\Value{y}=1 \right\} = \obj^{-1}(0) .
\ee
It follows from the computation of $d\obj/dt$
that the diagonal set $\Diag$ of $\Ygraph\times\Ygraph$ is repelling, 
and that the boundary cycle $\del\Finless$ is an attracting limit cycle. 
\endpf

The action of the vector field is to descend off of the ``fins''
of $\Cs$ onto the region $\Finless$, and then to circulate about
while pushing out to the boundary cycle $\del\Finless$, as in 
Figure~\ref{fig_field}. 

This example illustrates how one can use a relatively simple vector
field on the configuration space to construct a pattern which is 
free from collisions. In fact, one could use this circulating flow to 
regularize potential collisions between AGV's in a general graph setting 
by localizing the dynamics near a pairwise collision to those on a trivalent 
subgraph.

\section{Patterns and Vector Fields for Monotone Cycles}
\label{sec_5}\label{sec_4}
% ****************************************************

Optimization of patterns in the workspace is deeply entwined with
the geometry of the configuration space: in \cite{Ghr:geods},
it is shown that various Finsler structures on $\Cs$ can
be chosen to optimize total distance traveled or net time
elapsed.  The net result of this inquiry is that minimizing
Euclidean distance (in the product of the graph metric) on 
the cells of the configuration space yields locally optimal 
configuration sequences with respect to both distance traversed 
and elapsed time. In this section, we consider the problem of 
constructing vector fields which are tuned to trace out specific 
patterns of cooperative dynamics. We begin with a specification 
of a suitable language for describing patterns.

% ****************************************************
\subsection{A Grammar for Patterns}
% ****************************************************

The setting we envisage is as follows: the three ends of the graph
$\Ygraph$ are stations at which an AGV can perform some function. 
The AGV pair is required to execute an ordered sequence of 
functions, requiring an interleaved sequence of visitations. 
In order to proceed with vector field controls for cooperative
patterns, it is helpful to construct the appropriate symbolic language, 
as done in \S\ref{sec_2} for single AGV systems. 
Denote the pair of AGV states as 
$x$ and $y$ respectively. Also, denote the three docking stations 
as vertices $v_1$ through $v_3$ as in Figure~\ref{fig_Y}. 
The grammar $\Gs$ we use is defined as follows:
\begin{itemize}
\item 
	$\GramX{i}$: These represent configurations for 
	which the AGV $x$ is docked at the vertex $v_i$, $i=1..3$.
	The AGV $y$ is at an unspecified undocked position.
\item
	$\GramY{i}$: These represent configurations for 
	which the AGV $y$ is docked at the vertex $v_i$, $i=1..3$.
	The AGV $x$ is at an unspecified undocked position.
\item 
	$\GramXY{i}{j}$: These represent configurations for which
	the AGV $x$ is docked at vertex $v_i$ while the AGV $y$ 
	is simultaneously docked at the vertex $v_j$, $j\neq i$.
\end{itemize}
For example, the word $\GramX{1}\GramY{2}\GramXY{3}{2}$
executes a sequence in which the first AGV docks at Station $v_1$, then 
undocks while the second AGV docks at Station $v_2$. Finally, 
the AGV's simultaneously dock at Stations $v_3$ and $v_2$ 
respectively.

As we have assumed from the beginning, the one-dimensional 
nature of the graph-constraints precludes the presence of multiple
agents at a single docking station; hence, there are exactly twelve
symbols in the grammar $\Gs$. 
From this assumption, it follows that particular attention is to be 
paid to those trajectories which do not make excursions onto the 
``fins'' of the configuration space. It is obvious from the physical
nature of the problem that planning paths which involve traveling
on the fins is not a locally optimal trajectory with respect to minimizing
distance or elapsed time. Suffice to say that
we restrict attention for the moment to trajectories, and limit 
cycles for patterns in particular, which are constrained
to the region $\Finless\subset\Cs$.

\begin{figure}[htb]
\begin{center}
\psfragscanon
\psfrag{1}[l][r]{$\GramX{1}$}
\psfrag{2}[b][t]{$\GramXY{1}{2}$}
\psfrag{3}[b][t]{$\GramY{2}$}
\psfrag{4}[b][t]{$\GramXY{3}{2}$}
\psfrag{5}[b][t]{$\GramX{3}$}
\psfrag{6}[b][t]{$\GramXY{3}{1}$}
\psfrag{7}[r][l]{$\GramY{1}$}
\psfrag{8}[t][]{$\GramXY{2}{1}$}
\psfrag{9}[tr][]{$\GramX{2}$}
\psfrag{0}[t][]{$\GramXY{2}{3}$}
\psfrag{a}[tl][]{$\GramY{3}$}
\psfrag{b}[t][]{$\GramXY{1}{3}$}
	\includegraphics[width=2.0in\centerline]{{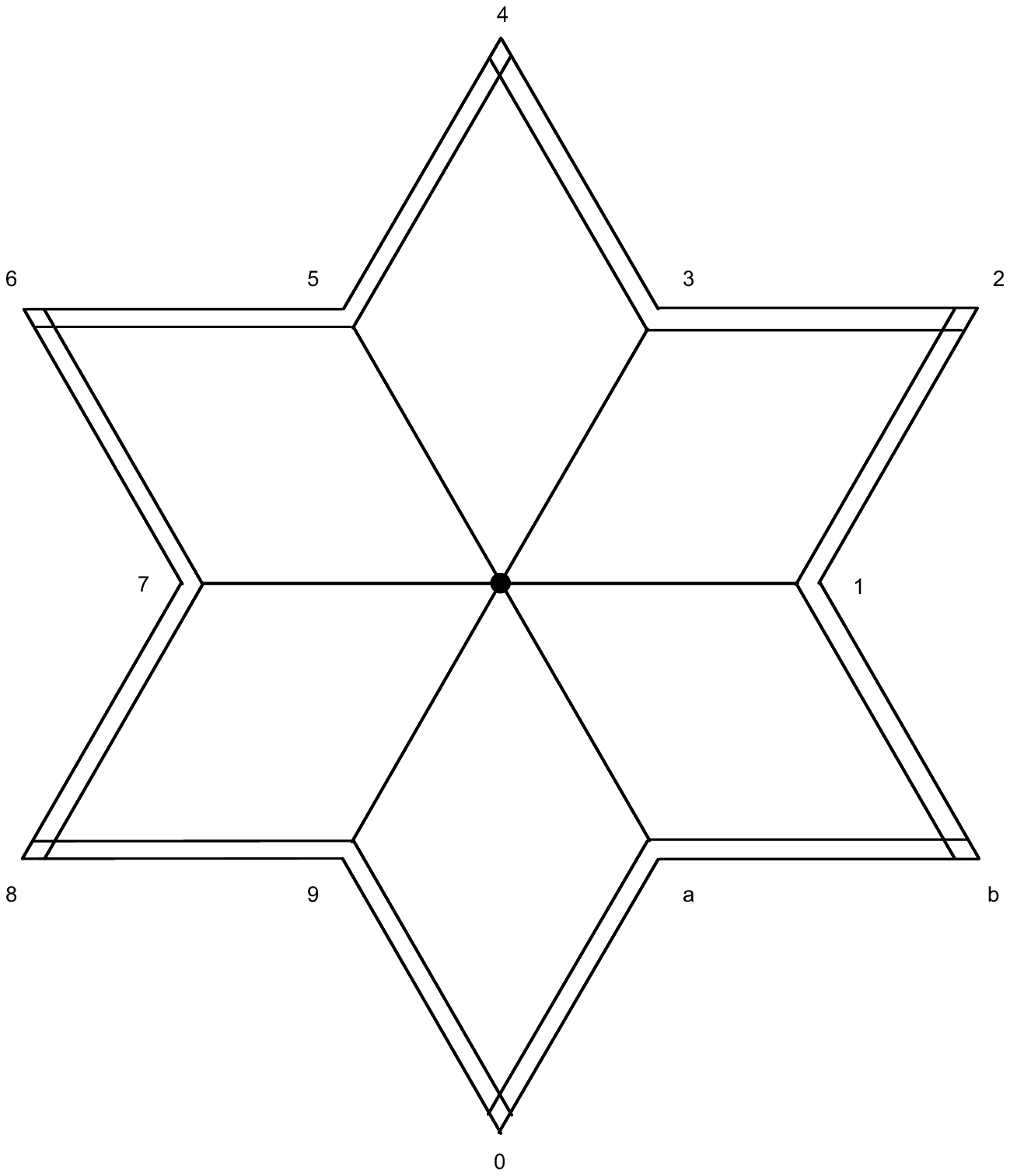}}
\end{center}

\caption{The six pairs of contiguous edges in $\del\Finless$ 
	each corresponds to a configuration
	where one AGV is docked at an extreme vertex
	of the graph. Outermost vertices of $\del\Finless$ 
	are points where both AGV's are docked. Labelling the
	edges $\GramX{i}$ and $\GramY{j}$ and the vertices
	$\GramXY{i}{j}$ yieldsd the cyclically 
	ordered grammar $\Gs$.} 
\label{fig_Zones}
\end{figure}

We identify each symbol with a region of the boundary of the 
unbranched portion of $\Cs$; namely, $\del\Finless$ is partitioned 
into twelve {\em docking zones} 
as in Figure~\ref{fig_Zones}. 
Note further that there is a cyclic ordering, $\lhd_\del$, on $\Gs$ 
induced by the orientation on the boundary of the disc along 
which the zones lie. By a cyclic ordering, we mean a way of 
determining whether a point $q$ lies between any ordered pair
of points $(p_1,p_2)$. 

%FIGURE ZONES
% =========================================
% =========================================

We proceed with the analysis of limit cycles on $\Cs$.
Consider the class of {\em pattern vector fields}, ${\cal X}_P$, 
on $\Cs$: for every $X\in{\cal X}_P$,
\begin{enumerate}
\item 
	$X$ defines a semiflow on $\Cs$ and a genuine flow off of the 
	non-manifold set of $\Cs$;
\item
	There is a unique limit cycle $\gamma$ which is attracting
	and which traces out a nonempty word in the grammar $\Gs$;
\item
	The diagonal set $\Diag$ is a repellor with respect to $X$;
\item
	There are no fixed invariant sets of $X$ which 
	attract a subset of positive measure save $\gamma$. 	
\end{enumerate}
Then, the class of {\em monotone vector fields}, ${\cal X}_{M}$,
is that subset of ${\cal X}_P$ for which the limit cycle, $\gamma$, 
lies in $\Finless$. A
word ${\bf w}$ composed of elements ${\bf w}=w_1w_2...w_n$ 
in the grammar $\Gs$ said to be {\em monotone} with respect 
to the cyclic ordering $\lhd_\del$ if $w_{i-1}\lhd w_i\lhd w_{i+1}$ for 
every $i$ (index operations all mod $n$).
The following result justifies our use of the term {\em monotone} 
in describing those limit cycles which lie on the disc.

\begin{thm}
Within the class of vector fields ${\cal X}_{M}$, the 
limit cycles trace out monotone words in the cyclically ordered 
grammar $(\Gs,\lhd_\del)$. 
\end{thm}

\begpf
Any limit cycle of the flow 
must be embedded (the curve does not intersect itself). 
After a small perturbation, one may assume that the boundary zone 
$\del\Finless$ is visited by $\gamma$ in a finite number of points,
$Q:=\gamma\cap\del\Finless$. There is a cyclic order $\lhd_t$, defined
via time with respect to the 
dynamics of the limit cycle: i.e., the order in which points are
visited by $\gamma$. This is contrasted with the induced cyclic ordering 
$\lhd_\del$ on the set $Q$ given by the orientation of the boundary curve 
$\del\Finless$, up to a choice between clockwise and 
counterclockwise. The theorem follows from showing that 
$\lhd_t=\lhd_\del$ up to a cyclic permutation and a choice of 
orientation of $\del\Finless$. 

\begin{figure}[htb]
\begin{center}
\psfragscanon
\psfrag{p}[br][]{$p$}
\psfrag{q}[bl][]{$q$}
\psfrag{s}[tr][]{$x$}
\psfrag{a}[][]{$\alpha$}{%\centerline{
	\includegraphics[width=1.4in]{{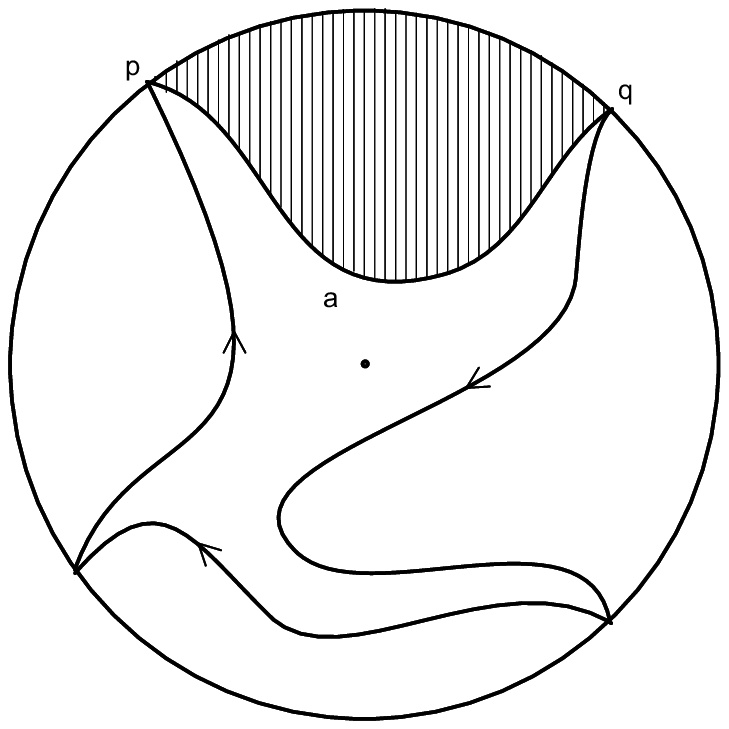}}
\hspace{0.15in}
	\includegraphics[width=1.4in]{{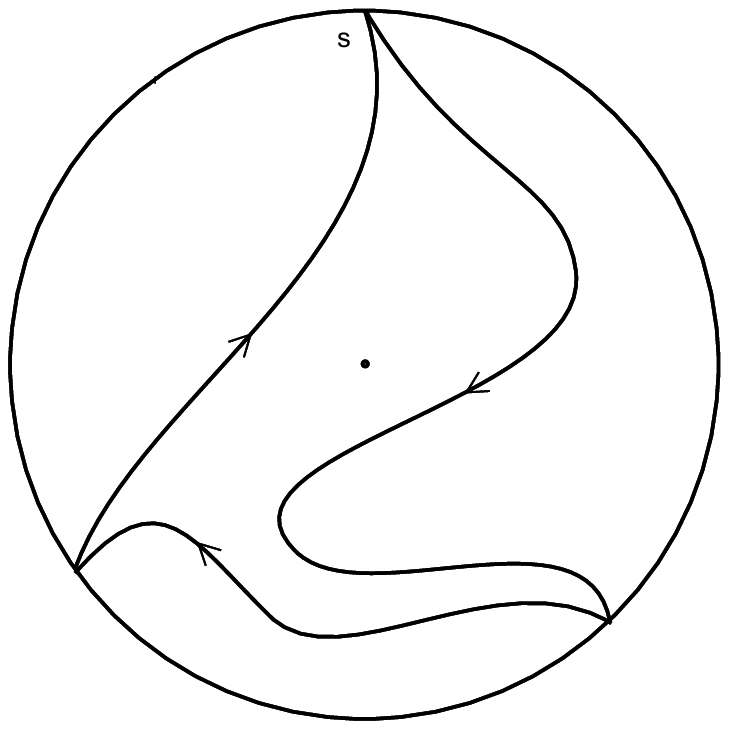}}}
\end{center}
\vspace{-0.1in}

\caption{The embedded arc $\alpha$ divides $\Finless$ (pictured as a smooth
	disc) into two discs, one of which is collapsed to a point $x$.} 
\label{fig_Collapse}
\end{figure}

Induct on $J$ the number of points in $Q$. For $J=1$, the theorem 
is trivially true, so assume that the orderings are equivalent for all 
embedded curves on a disc with less than $J$ boundary-intersections.
Consider two points $p, q\in Q$ which are consecutive in the $\lhd_t$ 
ordering. There is an embedded sub-arc $\alpha\subset\gamma$ which 
connects $p$ to $q$ within the interior of $\Finless$. By the Jordan Curve 
Theorem \cite{munkres}, $\alpha$ separates $\Finless$ into two topological 
discs; hence, $\gamma$ must lie entirely within the closure of one of 
these discs. Consider the open subdisc whose intersection with $\gamma$
is empty and collapse the closure of this disc to a point: this   
yields a modified curve $\tilde\gamma$ which 
is still embedded in a disc having $J-1$ intersections with the boundary,
as illustrated in Figure~\ref{fig_Collapse}.
By induction, the ordering $\lhd_t$ equals $\lhd_\del$ up
to orientation on this subdisc. Reinserting the distinct points 
$p$ and $q$ by ``blowing up'' the crushed disc does not change 
the ordering properties, since these were chosen to be adjacent.
\endpf

%FIGURE COLLAPSE
% =========================================
% =========================================

Hence, the ony admissible words in the grammar $\Gs$ are those 
which are monotone. 
It is, however, possible to realize many if not all of the non-monotone 
cycles as limit cycles for a semiflow on the {\em full} configuration space
$\Cs$; one must design the semiflow so as to utilize the fins for ``jumping''
over regions of $\Finless$ cut off by the limit cycle. Such vector 
fields quickly become very convoluted, even for relatively simple
non-monotone limit cycles.

%_!_!_!_!_!_!_!_!_!_!_!_!_!_!_!_!_!_!_!_!_!_!_!_!_!_!_!_!_!
\subsection{Isotopy Classes of Limit Cycles}
%_!_!_!_!_!_!_!_!_!_!_!_!_!_!_!_!_!_!_!_!_!_!_!_!_!_!_!_!_!

Given a limit cycle $\gamma$ which traces out a pattern by visiting
the boundary zone $\del\Finless$ in the ordered set 
$Q\subset\del\Finless$, one 
wants to know which other limit cycles minimize a given performance
functional while still visiting $Q$ in the proper sequence. The 
mathematical framework for dealing with this problem is the notion 
of {\em isotopy classes of curves}. 

Two subsets $A_0$ and $A_1$ of a set $B$ are said to be (ambiently) 
{\em isotopic rel $C$} (where $C\subset B$) if there exists a continuous 
1-parameter family of homeomorphisms, $f_t:B\ra B$ such that 
\begin{enumerate}
\item $f_0$ is the identity map on $B$;
\item $f_1(A_0)=A_1$; and
\item $\rest{f_t}{C}$ is the identity map on $C$ for all $t$. 
\end{enumerate}
As $t$ increases, $f_t$ deforms $B$, pushing  $A_0$ to $A_1$ 
without cutting or tearing the spaces and without disturbing $C$. 

There are two ways in which optimization questions relate
to isotopy classes of limit cycles: (1) Given an element of 
the grammar $\Gs$, 
in which isotopy class (rel the docking zones) of curves does an 
optimal limit cycle reside? (2) Within a given isotopy class
of cycles rel $Q$, which particular cycle is optimal?

For a monotone limit cycle on $\Finless$, the above 
question (1) focuses on the location of the cycle with respect to 
the central point $(0,0)$, which is deleted from the disc $\Finless$. 
It is a standard fact from planar topology that every curve in 
the punctured disc has a well-defined {\em winding number}, which 
measures how many times the cycle goes about the origin, and, 
furthermore, that this number is either -1, 0, or 1 if the cycle is 
an embedded curve. This winding number
determines the isotopy class of the curve in $\Finless$. 
Hence, the problem presents itself: given an element of the 
grammar $\Gs$, which isotopy class rel the docking zones 
is optimal (with respect to any/all of the functionals defined)?
Is the winding number zero or nonzero?\footnote{The difference between 
+1 and -1 is the orientation of time.} 

To briefly address this question, we define the {\em gap angles} 
associated to a limit cycle. Given the docking zones 
$Q=\{q_1, q_2, \ldots, q_J\}$ ordered with respect to time, 
we define the gap angles to be the successive differences in 
the angular coordinates of the $q_j$: thus $ga_j:=P(q_{j+1})-P(q_j)$,
where $P$ denotes projection of points in $\Finless$ onto their 
angular coordinates and subtraction is performed with respect to the
orientation on $\partial\Finless$. 

For simplicity, we consider the optimization--isotopy problem 
in the case of a discrete cost functional $\DWork$, defined
to be the intersection number of the path with the branch
locus of $\Cs$ --- i.e., the number of times an AGV occupies
the central vertex (the shared resource in the problem). 
Similar arguments are possible for other natural  
performance metrics \cite{Ghr:geods}.

\begin{prop}
There is a $\DWork$-minimizing embedded monotone cycle on $\Finless$ 
(rel a given docking zone $Q$) having  
winding number zero with respect to the origin if 
there is a gap angle greater than $\pi$. Conversely, if 
there are no gap angles greater than $\pi$, then there is 
a $\DWork$-minimizing embedded cycle of index $\pm 1$.
\end{prop}

 \begpf
 Let $Q$ consist of the points $\{q_j\}_1^J$ on the boundary circle. 
 The gap angles $\{ga_j\}_1^{J}$ are the differences of the 
 angles between the points $q_j$ and $q_{j+1}$ (indices mod $J$). Since 
 $\sum_j{ga}_j=2\pi$, there can be at most one gap angle greater than $\pi$. 
 To simplify the problem, use a 1-parameter family $P_t$ of 
 maps from the identity $P_0$ to the projection $P=P_1$ 
 which deforms $\Finless$ to the boundary circle $S:=\del\Finless$ 
 by projecting continuously 
 along radial lines. The index of a curve on $\Finless$ is invariant 
 under this deformation, as is the function $\DWork$. 

 Denote by $\gamma_j$, the subarc of $\gamma$ between points $q_{j}$
 and $q_{j+1}$ (all indices mod $J$). Denote by $\alpha_j$ the subarc 
 of the boundary $S$ between points $q_j$ and $q_{j+1}$, where the 
 arc is chosen to subtend the gap angle $ga_j$. Since the boundary
 curve $S=\cup_j\alpha_j$ is a curve of index $\pm 1$, the arcs 
 $\gamma_j$ and $\alpha_j$ are isotopic in $\Finless$ rel their 
 endpoints for all $j$ if and only if $\gamma$ is a curve of index $\pm 1$. 

 Assume first that there is a gap angle $ga_j>\pi$ with $\gamma$ 
 an index $\pm 1$ curve on $S$ which intersects the branch angles 
 $\Theta=\{n\pi/3 : n\in Z\}$ in a minimal number
 of points among all other closed curves on $S$ which visit the points
 $Q$ in the specified order. 
 It follows that the arc $P(\ga_j)$ subtends an angle greater than
 $\pi$ and thus increments $\DWork$ by at least three. 
 One may replace $\ga_j$ by a curve $\ga_j'$ which substitutes for the 
 arc $\ga_j$ one which wraps around `the other way' monotonically.
 This changes the index of $\ga$ from nonzero to zero, since the 
 arc $\ga_j'$ is no longer isotopic to $\alpha_j$. Also, 
 it is clear that this either decreases the number of intersections 
 with $\Theta$ or leaves this number unchanged. 

 We must show that the replacement arc $\ga_j'$ can be 
 chosen in such a way that is does not intersect the remainder of 
 $\ga$. However, since $\ga$ is a curve of index $\pm 1$, we may 
 isotope each arc $\ga_i$ to the boundary curve $\alpha_i$ without 
 changing the value of $\DWork$. Thus, we may remove $\ga_j$ and 
 replace it with the curve which is, say, a geodesic (in the natural
 metric geometry)  
 from $q_j$ to $q_{j+1}$. As this curve does not approach the 
 boundary $S$ apart from its ends, the new curve $\ga'$ is an 
 embedded curve of index zero without an increase in $\DWork$.

 Assume now, on the contrary, that $\gamma$ is a $\DWork$ minimizer
 of index zero which has all gap angles strictly less than $\pi$. 
 Then each arc from $\ga_i$ must intersect the branch set
 $\Theta$ in at most three components, since, otherwise, the 
 subtended arc would be in excess of $4\pi/3$.
 In the case where there exists an arc with exactly three 
 intersections with the branch set, this arc may be replaced 
 by an arc which goes around the singularity in the other direction 
 without changing the number of intersections with the branch set
 (since there are a total of six branch lines); however, the 
 index of the curve is toggled between zero and nonzero. 

 The final case is that in which each arc intersects the branch 
 set in at most two places. However, since $\ga$ is a curve of index
 zero, some arc $\ga_j$ must not be isotopic to $\alpha_j$. Hence, 
 the projection deformations $P_t$ must push $\ga_j$ to a curve in 
 the boundary which $S$ whose subtended gap 
 angle is $2\pi - ga_j > \pi$. Thus, $\gamma_j$ intersects the branch 
 set in at least three places, yielding a contradiction. Replacing 
 $\ga_j$ by the appropriate arc which is isotopic to $\alpha_j$ 
 yields a $\DWork$-minimal cycle of nonzero index.
 \endpf

% ****************************************************
\subsection{Tuning Cycles}
% ****************************************************

In order to proceed with the construction of vector fields which 
realize monotone cycles, we work with vector fields 
on the smooth unit disc in $R^2$ and map these to the annular 
region $\Finless$ of the configuration space via the push-forward induced
by the natural homeomorphism. It will be convenient to keep track 
of which ``wedge'' of the annular region a point $(r,\theta)$ is. 
To do so, we introduce a parity function 
\be\label{eq_Parity}
P(\theta) := (-1)^{\left\{ \floor{3\theta/\pi}+
			  \floor{6\theta/\pi} \right\}}
,\ee
where $\floor{t}$ is the integer-valued floor function.
Recall the notation for the intrinsic coordinates for a point 
$x$ on the graph $\Ygraph$: $x=\Value{x}\hat{e}_{\Index{x}}$, 
where $\Value{x}\in[0,1]$ is the distance from $x$ to the 
central vertex, and $\hat{e}_{\Index{x}}$ is the unit tangent vector
pointing along the direction of the $\Index{x}$-edge. Here
the index, $\Index{x}$ is an integer (defined modulo 3) and 
will be undefined in the case when $\Value{x}=0$, i.e., $x$ 
is at the central vertex. 

\begin{lem}
The following is a piecewise-linear homeomorphism from the punctured 
unit disc in $R^2$ to the subset $\Finless$: 
$F(r,\theta) = (x, y)$ where,
\be \label{eq_Homeo}
\begin{array}{rl}
	& \Value{x} = \left\{\begin{array}{cl}
		r & \Prt{\theta}=+1 \\
		r\abs{\cot\frac{\ds 3}{\ds 2}\theta} & \Prt{\theta}=-1
		\end{array}\right.   \\
	& \Value{y} =    \left\{\begin{array}{cl}
		r\abs{\tan\frac{\ds 3}{\ds 2}\theta} & \Prt{\theta}=+1 \\
		r & \Prt{\theta}=-1 
		\end{array}\right.   \\
	& \Index{x} = \floor{-\frac{\ds 3}{\ds 2\pi}(\theta-\pi)}   \\
	& \Index{y} = \floor{-\frac{\ds 3}{\ds 2\pi}\theta}   
\end{array} .
\ee
The inverse of this homeomorphism is given by
$F^{-1}(x,y) =  (r,\theta)$, where,
\be \label{eq_Inverse}
\begin{array}{l}
	\theta = \left\{\begin{array}{cl}
		\frac{\ds 2}{\ds 3}\tan^{-1}\frac{\ds\Value{y}}{\ds\Value{x}}
		- \frac{\ds 2\pi}{\ds 3}(\Index{y}+1) 
		& \begin{array}{c}
			\Index{y}=\Index{x}+1 \\
			{\mbox{ or }} \Value{x}=0 
		     \end{array}  \\
		 -\frac{\ds 2}{\ds 3}\tan^{-1}\frac{\ds\Value{y}}{\ds\Value{x}}
		- \frac{\ds 2\pi}{\ds 3}(\Index{x}-1) 
		& \begin{array}{c}
			\Index{x}=\Index{y}+1 \\
			{\mbox{ or }} \Value{y}=0
		     \end{array}  
		\end{array}\right.   \\
	r  = \left\{\begin{array}{cl}
			\Value{x} & \Prt{\theta} = +1 \\
			\Value{y} & \Prt{\theta} = -1
		\end{array}\right.
\end{array} . \ee
\end{lem}
Note that all $\theta$ values are defined modulo $2\pi$ and 
all index values are integers defined modulo $3$.

\begpf
Begin by working on the region $\Finless_{1,2}\subset
\Finless$ where $\Index{x}=1$ and $\Index{y}=2$. As noted earlier, this 
subspace is isometric to the positive unit square in $R^2$ with 
the origin removed. We need to map this to the subset 
$\{(r,\theta): r\in(0,1], \theta\in[0,\pi/3]\}$. The simplest such 
homeomorphism is to first shrink along radial lines, leaving the 
angle invariant; hence
\be \label{eq_r}
	r = \left\{ \begin{array}{cl}
		\Value{x} &: \Value{x}\leq\Value{y} \\
		\Value{y} &: \Value{y}\leq\Value{x}
	\end{array}
\right.\ee
Next, we squeeze the quarter-circle into a sixth of a circle by 
multiplying the angle by $2/3$, leaving the radial coordinate 
invariant:
\be\label{eq_theta}
	\theta = \frac{\ds 2}{\ds 3}\tan^{-1}\frac{\ds\Value{y}}{\ds\Value{x}} .
\ee
This gives the basic form of $F^{-1}$ as per Equation~(\ref{eq_Inverse}).
To extend this to the remainder of $\Finless$, it is necessary to
carefully keep track of $\Index{x}$ and $\Index{y}$ and subtract
off the appropriate angle from the computation of $\theta$. Also, 
the condition of $\Value{x}\leq\Value{y}$, etc., in 
Equation~(\ref{eq_r}) is incorrect 
on other domains of $\Finless$, since the inequalities flip as one 
traverses from square to square: the parity function $\Prt{\theta}$ 
keeps track of which ``wedge'' one is working on. 

To determine $F$ from $F^{-1}$ is a tedious but unenlightening
calculation, made more unpleasant by the various indices to be 
kept track of. Briefly, given $r$ and $\theta$ on the first sixth 
of the unit disc, one knows from Equation~(\ref{eq_r}) that either
$\Value{x}=r$ or $\Value{y}=r$, depending on whether $\theta$
is above or below $\pi/4$. To solve for the other magnitude, one 
inverts Equation~(\ref{eq_theta}) to obtain $\Value{y} = 
r\abs{\tan\frac{\ds 3}{\ds 2}\theta}$ or $\Value{x} = 
r\abs{\cot\frac{\ds 3}{\ds 2}\theta}$
respectively. To generalize this to the other 
$\Finless_{i,j}$ domains of $\Finless$, 
it is necessary to take absolute values and to use the parity 
function $\Prt{\theta}$ as before. Finally, the computation of the
index is obtainable from the combinatorics of the coordinate 
system as illustrated in Figure~\ref{fig_Coords}.
\endpf

Hence, by taking the push-forward of a vector field $X=(\dot{r} ,
\dot{\theta})$ with respect to $F$, one obtains the piecewise smooth 
vector field, 
\be \label{eq_Pushforward}
	\left\{\begin{array}{rc}
	\left( \begin{array}{l}
		\dot{\Value{x}} = \dot{r} \\
		\dot{\Value{y}} = \dot{r}\abs{\tan(\frac{\ds 3}{\ds 2}\theta)}
			+\frac{\ds 3}{\ds 2}r\dot{\theta}\sec^2(\frac{\ds 3}{\ds 2}\theta)
	\end{array}\right) & \Prt{\theta}=+1 \\
	\left( \begin{array}{l}
		\dot{\Value{x}} = \dot{r}\abs{\cot(\frac{\ds 3}{\ds 2}\theta)}
			+\frac{\ds 3}{\ds 2}r\dot{\theta}\csc^2(\frac{\ds 3}{\ds 2}\theta) \\
		\dot{\Value{y}} = \dot{r}
	\end{array}\right) & \Prt{\theta}=-1 
\end{array}\right. ,\ee
which simplifies to:
\be \label{eq_Pushforward2}
	\left\{\begin{array}{rc}
	\left( \begin{array}{l}
		\dot{\Value{x}} = \dot{r} \\
		\dot{\Value{y}} = \dot{r}\frac{\ds \Value{y}}{\ds \Value{x}}
			+\frac{\ds 3}{\ds 2}\dot{\theta}\frac{\ds \Value{x}}
			{\ds 1+\left(\frac{\ds \Value{y}}{\ds \Value{x}}\right)^2} 
	\end{array}\right) & \Prt{\theta}=+1 \\
	\left( \begin{array}{l}
		\dot{\Value{x}} = \dot{r}\frac{\ds\Value{x}}{\ds\Value{y}}
			+\frac{\ds 3}{\ds 2}\dot{\theta}\frac{\ds \Value{y}}
			{\ds 1+\left(\frac{\ds\Value{x}}{\ds\Value{y}}\right)^2}  \\
		\dot{\Value{y}} = \dot{r}
	\end{array}\right) & \Prt{\theta}=-1 
\end{array}\right. .\ee

Given a simple closed curve $\gamma$ in $R^2$ which has nonzero
winding number with respect to the origin, $\gamma$ may be 
parametrized as $\{(r,\theta) : r=f(\theta)\}$ for some periodic
positive function $f$. To construct a vector field on $R^2$ 
whose limit sets consist of the origin as a source and $\gamma$ as an 
attracting limit cycle, it suffices to take the push-forward of 
the vector field 
$ 	\dot{r} = r(1-r) \hspace{0.3in} \dot{\theta}=\omega $
under the planar homeomorphism 
$	\phi:(r,\theta)\mapsto (f(\theta)r,\theta), $
which rescales linearly in the angular component. The calculations follow: 
\be\label{eq_PushCycle}
\begin{array}{rl}
\phi_*\left(\begin{array}{c} \dot{r} \\ \dot{\theta}\end{array}\right)
& = \left. D\phi
	\left(\begin{array}{c}\dot{r} \\ \dot{\theta}\end{array}\right)
	\right\vert_{r\mapsto\frac{\ds r}{\ds f}} 
 = \left.\left[\begin{array}{cc}
	f & rf' \\ 0 & 1 \end{array}\right]
	\left(\begin{array}{c}r(1-r) \\ \omega \end{array}\right)
	\right\vert_{r\mapsto\frac{\ds r}{\ds f}} \\
& = \left(\begin{array}{c}
	r\left(1-\frac{\ds r-f'\omega}{\ds f}\right) \\
	\omega	\end{array}\right)
\end{array} .\ee
Hence, given $f(\theta)$, we may tune a vector field to trace out 
the desired limit cycle and then use Equations~(\ref{eq_Homeo})
and (\ref{eq_Inverse}) to map it into intrinsic coordinates.

% ****************************************************
\subsection{Optimal Chords within a Hybrid Controller} 
% ****************************************************

To design optimal cycles with winding number zero, then, we 
turn to constructing customized portions of limit cycles, or 
{\em chords} which can be pieced together via a state-actuated 
hybrid controller, much as in \S\ref{sec_2}. 
In other words, instead of building a simple fixed
vector field with a limit cycle, we will use a set of vector
fields which vary discretely in time and which may be pieced 
together so as to tune a limit cycle to the desired specifications. 
There is nothing in this construction which relies on the index-zero
property and thus these chords can be used to generate all 
monotone limit cycles on $\Cs$.

Let $\Goalword$ denote a word representing a desired monotone 
limit cycle on the configurations space $\Cs$. Choose points 
$\{q_i\}$ on the boundary of $\Finless$ which correspond to the 
docking zones for the cycle given by $\Goalword$. Choose 
arcs $\alpha_i$ on $\Finless$ which connect $q_i$ to $q_{i+1}$
(using cyclic index notation). The arcs $\alpha_i$ are assumed 
given in the intrinsic coordinates on $\Finless$, as would be the
case if one were determining a length-minimizing curve. 

In the case where the limit cycle $\alpha:=\cup_i\alpha_i$ is an 
embedded curve of nonzero index, the procedure of the previous subsection
determines a vector field $X_\alpha$ on $\Cs$ which realizes $\alpha$ as an 
attracting limit cycle with the appropriate dynamics on the 
complementary region. Recall: one translates $\alpha$ to a curve 
on the disc model via the homeomorphism of Equation~(\ref{eq_Inverse}).
Then, representing the limit cycle $\alpha$ as a function 
$f_\alpha(\theta)$, one takes the vector field of 
Equation~(\ref{eq_PushCycle}) and, if desired, takes the 
image of this vector field under Equation~(\ref{eq_Pushforward2}).  

If, however, this is not the case, consider
the arc $\alpha_j$ for a fixed $j$ and construct an index $\pm 1$ 
cycle $\beta^j=\cup_i\beta_i^j$ which has docking zones $\{q_i\}$ 
such that $\beta_j^j=\alpha_j$. Then the vector field $X^j$ 
as constructed above has $\beta$ as an attracting 
limit cycle. Denote by $\Phi^j$ the Lyapunov function which measures 
proximity to $\beta$: $\Phi^j(p) := \norm{p-\beta^j}$ 
(with distance measured in say the product metric on $\Cs$). 
Then, consider the modified Lyapunov function 
%\begin{equation}
%\label{eq_ModLyap}
$	\Psi^j(p) := \Phi^j(p) + \norm{p-q_{j+1}},$
%\end{equation}
which measures the distance to the endpoint of the arc $\beta_j^j$
in addition to the proximity to $\beta^j$. 

Repeat this procedure for each $j$, yielding the vector fields 
$\{X^j\}$ which attract respectively to limit cycles $\beta^j$. 
It follows that $X^j$ prepares $X^{j+1}$ 
since the goal point of $X^j$, $q_{j+1}$ lies on the attracting 
set of $X^{j+1}$. The Lyapunov functions $\{\Psi^j\}$ serve as 
a set of funnels which channel the orbit into the sequence of 
arcs $\alpha_j$, forming $\alpha$. One scales the $\Psi^j$ so that 
a $\Psi^j<\epsilon$ event triggers the switching in the hybrid 
controller from $X^j$ to $X^{j+1}$:
\begin{equation}
\label{eq_Controller}
X := \left\{ \begin{array}{ccl}
X^1 &:& \Phi^j>\epsilon \, \forall\, j \\
X^j &:& \Phi^j<\epsilon {\mbox{ and }} \Psi^j>\eps
\end{array}\right.
\end{equation} 

By construction, the hybrid controller (\ref{eq_Controller}) 
realizes a limit cycle within $\epsilon$ of 
$\alpha$ as the attracting set.

\section{Future Directions}
\label{sec_6}

A point of primary concern is the adaptability of the global 
topological approach to systems which increase in complexity, 
either through more intricate graphs or through increased 
numbers of AGV's. The latter is of greater difficulty than the 
former, since the dimension of the resulting configuration 
space is equal to the number of AGV's. Hence, no matter how 
simple the underlying graph is, a system with ten independent 
AGV's will require a dynamical controller on a (topologically 
complicated) ten-dimensional space: a formidable problem 
both from the topological, dynamical, and computational 
viewpoints. 

However, there are some approaches which may facilitate working 
with such spaces. Consider the model space $\Cs$ with which this 
paper is concerned: although a two-dimensional space, $\Cs$ 
can be realized as the product of a graph (a circle with six 
radial edges attached) with the interval $(0,1]$. In fact, if we consider 
the circulating flow of Equations~(\ref{eq_Dance1})-(\ref{eq_Dance3}), 
one can view this as a product field of a semiflow on the graph 
(which ``circulates'') with a vector field on the factor $(0,1]$ 
(which ``pushes out'' to the boundary). 

A similar approach is feasible for arbitrary graphs. 
The following result has recently been proven \cite{Ghr:birman}:
\begin{thm}
Given a graph $\Gamma$, the configuration space of $N$ distinct 
points on $\Gamma$ can be deformation retracted to a subcomplex 
whose dimension is bounded above by the number of vertices of 
$\Gamma$ of valency greater than two.\footnote{We have since
learned of two others who have independently proved this result: 
\cite{MK99}, having learned of these spaces fron \cite{GK:japan}; 
and \cite{AA99}, who discovered these spaces while working on 
the topology of Brownian motion on graphs.}
\end{thm}

This theorem implies the existence of low-dimensional {\em spines} 
which carry all of the topology of the configuration space. 
For example, the above theorem implies that the configuration 
space of $N$ points on the Y-graph can be continuously deformed 
to a one-dimensional graph, regardless of the size of $N$. Since 
the full space can be deformation retracted onto the spine, 
a vector field defined on the spine can be pulled 
back continuously to the full configuration space, thus opening 
up the possibility of reducing the control problem to that 
on a simpler space. Addtional results about the topology of
configuration spaces on graphs may yield computationally
tractible means of dealing with complex path planning: for
example, having a presentation for the fundamental 
group of a configuration space of a graph in terms of 
a suitably simple set of cycles would be extremely well-suited
to a hybrid control algorithm based on ``localized'' 
vector fields supported on small portions of the full 
configuration space.

The optimization problem is another avenue for inquiry. The 
fact that a dynamical approach allows for increased density
of AGV's on a graph (as compared with blocking-zone strategies)
would indicate an increased efficiency with respect to, say,
elapsed time-of-flight. However, a more careful investigation
of the tuning of optimal cycles is warranted.

We believe that the benefits associated with using the 
full configuration space to tune optimal dynamical cycles 
justifies a careful exploration of these challenging spaces.

\appendix

%\input{app_a.tex}
%%%%%%%%%%%%%%%%%%%%%%%%%%%%%%%%%%%%%%%%%
\section{The Topology and Dynamics of Graphs}
%%%%%%%%%%%%%%%%%%%%%%%%%%%%%%%%%%%%%%%%%

In this appendix, we provide a careful basis for the 
use of vector fields on configuration spaces of graphs.
In the setting of manifolds, all of the constructions 
used in this paper are entirely natural and well-defined.
However, on spaces like $\Cs$, the most fundamental of
notions (like the Existence and Uniqueness Theorems for
ODEs) are not in general valid.

We begin by defining vector fields on graphs.
For present purposes, we find it convenient to work with an intrinsic
formulation (i.e., directly in the graph rather than via an embedding in a
Euclidean space) of these objects. To this end, denote by
$v$ a vertex with $K$ incident edges $\{e_i\}_1^K$, and by
$\{X_i\}_1^K$ a collection of nonsingular vector fields locally
defined on a neighborhood of the endpoint of each $e_i$ (homeomorphic
to $[0,1)$). 

\begin{lem}\label{lem_Semiflow}
A set of nonsingular vector fields $\{X_i\}$ on the 
local edge set of a graph $\Gamma$ 
generates a well-defined semiflow on $\Gamma$ if
\begin{enumerate}
\item Each edge field $X_i$ generates a well-defined local semiflow
on (0,1); and
\item The magnitude of the endpoint vectors $\norm{X_i(0)}$ (taken with
respect to the attaching homeomorphisms) are all identical; and
\item Among the {\em signs} of the endpoint vectors $X_i(0)$ 
(either positive if pointing into $[0,\epsilon)$ or negative if 
pointing out) there is a single positive sign.
\end{enumerate}
\end{lem}

\begpf
Since the vector field is well-defined away from the vertex, 
it is only necessary to have the magnitudes $\norm{X_i(0)}$ agree
in order to have a well-defined function $\norm{X}$ on $\Gamma$.
In order to make this a well-defined field of directions, we
must also consider in which direction the vector is pointing.
Again, this is determined off of the vertex by (1). Condition
(3) means that at the vertex, there is a unique direction 
along which the vector field is pointing out: all other
edges point in. Hence, the direction field, as well as the magnitude
field, is well-defined.

The semiflow property follows naturally from this. Assume that 
the $N^{th}$ edge of $\Gamma$ has the positive sign. Then, given an initial
point $x\in\Gamma$, if $x\in e_N$, then the orbit of $x$ under the 
local field $X_N$ remains in $e_N$ and is well-defined. If 
$x\in e_j$ for some $j\neq N$, then the union of the edges
$e_j\cup e_N$ is a manifold homeomorphic to $R$ on which the
vector fields $X_j$ and $X_N$ combine to yield a well-defined
vector field, since the directions are ``opposite.'' 
As we are now on a manifold, the standard Existence
Theorem implies that $x$ has a forward orbit (which passes through the
vertex and continues into $e_N$). Thus every point on $\Gamma$ 
has a well-defined forward orbit.
\endpf

In the case where the vector fields have singularities, it is a 
simpler matter. If the singularities are not at the vertex, then 
there is no difference. If there is a singularity at the vertex, 
then condition (3) in Lemma~\ref{lem_Semiflow} is void --- 
all such vector fields are well-defined.

In order to extend these results to the configuration space of 
this paper, consider the space $\Cs=\Ygraph\times\Ygraph-\Diag$ and 
let $(x,y)\in\Cs$ denote a point on the branch set of $\Cs$. 
Because of the structure of $\Ygraph$ and the fact that the
diagonal points are deleted, it follows that at most one AGV
may occupy a non-manifold point of $\Ygraph$. Hence, a neighborhood
of $(x,y)$ in $\Cs$ has a natural product structure
$N\cong\Ygraph\times R$. Let $P:N\rightarrow\Ygraph$ denote 
projection onto the first factor. 

\begin{lem}
\label{lem_Semiflow2}
A nonsingular vector field $X$ on the individual cells of $\Cs$ 
generates a well-defined semiflow if (1) the projection of the local  
vector fields onto the graph factor, $P_*(\rest{X}{\{x\}\times\Ygraph})$, 
satisfies Lemma~\ref{lem_Semiflow} for each point $x$ in the branch
set of $\Cs$; and (2) the projections of the vector fields on the branch
set to the $R$-factor are equal up to the attaching maps.
\end{lem}

\begpf
Off of the branch set, the space is a manifold and hence the 
vector field gives a well-defined flow. If $p$ is a point on the 
branch line, condition (2) implieds that the vector field is well-defined 
with respect to the attaching maps and the net effect in the $R$-factor
is a drift in this direction. In the graph factor, condition (1) and
the proof of Lemma~\ref{lem_Semiflow} implies that there is a unique
forward orbit through $p$.
\endpf

Heuristically, this condition means that, as in the case of 
a graph, the vector field must point ``in'' on all but one
sheet of the configuration space in order to have well-defined
orbits. We may thus lift the criteria of Lemma~\ref{lem_Semiflow}
to the product configuration space. All of the vector fields
in this paper are so constructed.

\bibliography{}

\end{document}

%% file: setups.tex
    \newcommand{\norm}[1]{ \left\| #1 \right\| }

    \def \begpf{ {\bf Proof: } }
		%\begin{description} \item[{\bf Proof:\hspace{1em} }]  }
    \def \endpf{ \hfill $\diamond$ \vspace{0.1in} }

    \def \begrm{ \begin{description} \item[{\bf Remark:\hspace{2.0em} }]  }
    \def \endrm{ % \begin{ clearflushleft} $\diamond $ \end{flushleft}
                  \end{description} }

    \def \be { \begin{equation}   }
    \def \ee { \end{equation}  }

    \newcommand{ \choicetwo }[5] {
        #1 \ldf
                         \left\{ \begin{array}{c  l }
                           #2  & \st #3     \\
                           #4  & \st #5
                          \end{array} \right.  }

    \newcommand{\beq}[2]{ \be \label{eq.#1} #2 \ee }

    \def \bea { \begin{equation}   \begin{array}{ c l}
               }
    \def \eea { \end{array} \end{equation}  }

    \newcommand{\beaq}[2]{ \bea \label{eq.#1} #2 \eea }

    \def \bma { \[ \begin{array}{ c l}
               }
    \def \ema {    \end{array}  \] }

    \newcommand{\refig}[1]{ Figure~\ref{fig.#1}}
    \newcommand{\req}[1]{(\ref{eq.#1})}

% First Label Math Statements Properly:
%   Propositions and Lemmas  within sections numbered the same:
%  ________________________________________
	\newtheorem{thm}{Theorem}%[section]
% _________________________________________
    \newtheorem{prop}{Proposition}%[section]
    \newtheorem{lem}[prop]{Lemma}

%   Corollaries separately,  within sections:
    \newtheorem{cor}[prop] {Corollary} %[section]

\newcommand{\lthm}[2]{\begin{thm}\label{thm.#1} #2 \end{thm}}
    \newcommand{\rethm}[1]{Theorem~\ref{thm.#1}}

\newcommand{\lprop}[2]{\begin{prop}\label{prop.#1} #2 \end{prop}}
    \newcommand{\reprop}[1]{Proposition~\ref{prop.#1}}

\newcommand{\llem}[2]{\begin{lem}\label{lem.#1} #2 \end{lem}}
    \newcommand{\relem}[1]{Lemma~\ref{lem.#1}}

\newcommand{\lcor}[2]{\begin{cor}\label{cor.#1} #2 \end{cor}}
    \newcommand{\recor}[1]{Corollary~\ref{cor.#1}}

 \newcommand{ \matwo }[4] {
                         \left[ \begin{array}{c c }
                           #1  &  #2     \\
                           #3  &  #4
                          \end{array} \right]  }

\newcommand{ \prl } [1] {  \left( #1 \right) }
\newcommand{ \brl } [1] {  \left[ #1 \right] }
\newcommand{ \trp } [1] { #1^{\mathtt T} }

   \newcommand{ \vctwo }[2] {
                         \left[ \begin{array}{c }
                           #1        \\
                           #2
                          \end{array} \right]  }

  \newcommand{ \inv }[1] { {#1}^{-1} }

\def \ldf { {:=} }
\def \st { {:} }

\def \As { { \mathcal A } }
\def \Bs { { \mathcal B } }
\def \Cs { { \mathcal C } }
\def \Ds { { \mathcal D } }
\def \Es { { \mathcal E } }
\def \Gs { { \mathcal G } }
\def \Hs { { \mathcal H } }
\def \Ls { { \mathcal L } }
\def \Ms { { \mathcal M } }
\def \Ns { { \mathcal N } }
\def \Os { { \mathcal O } }
\def \Ts { { \mathcal T } }
\def \Vs { { \mathcal V } }
\def \Xs  { { \mathcal X } }
\def \Ys  { { \mathcal Y } }
\def \Graph { {\Gamma} }
\def \Goal{ {\mbox{{\sf g}}} }               %{ \mathsf g }
\def \goal{ {g} }

\def \real { {R} } %{\mbox{\rm I} \hspace{-.03in} {\bf R}} }	

\def \cube { { \xi } }
\def \EPF { {X} }
\def \Diag { { \Delta } }
\def \Ygraph { { \Upsilon } }
\def \Lsgraph {{ \Lambda }}
\def  \obj { {  \Phi } }
\def  \ener { {  \eta } }
\def  \dener { {  \dot{\eta} } }
\def \Finless { { \Ds } }
\def \del { { \partial } }
\def \ra { { \rightarrow } }
\def \DWork { { {\bf W}_d }}
\def \Pty { { {\cal P} }}
\def \lhd { { < } }
\newcommand{\Value}[1]{\vert #1\vert}                      
\newcommand{\abs}[1]{\left\vert #1\right\vert}
\newcommand{\Index}[1]{\iota({#1})}
\renewcommand{\arg}[2]{{\mbox{arg}}(#1,#2)}
\newcommand{\Prt}[1]{{\cal P}(#1)}
\newcommand{\ceil}[1]{\left\lceil #1 \right\rceil}
\newcommand{\floor}[1]{\left\lfloor #1 \right\rfloor}
\newcommand{\GramX}[1]{{\bf (x #1)}}
\newcommand{\GramY}[1]{{\bf (y #1)}}
\newcommand{\GramXY}[2]{{\bf (x #1 y #2)}}
\newcommand{\rest}[2]{\left. #1\right\vert_{#2}}		
\newcommand{\GoalWord}{{\mbox{\sf G}}}
\newcommand{\Goalword}{{\mbox{\sf G}}}
\newcommand{\ga}{\gamma}
\newcommand{\eps}{\epsilon}
\newcommand{\ds}{\displaystyle}